\documentclass[final,3p,times,twocolumn]{elsarticle}

\usepackage{blindtext, graphicx, amsmath, algorithm, algpseudocode, pifont, algcompatible, comment, layout, amsthm, amssymb}
\usepackage{enumitem}   
\usepackage{eso-pic}
\usepackage{booktabs}
\usepackage{float}
\usepackage[utf8]{inputenc}
\usepackage{enumitem}

\usepackage[utf8]{inputenc}
\usepackage[english]{babel}
\usepackage{hyperref} 
\hypersetup{ colorlinks=true, linkcolor=black, filecolor=black, urlcolor=cyan, }

\usepackage{caption}
\usepackage{amsmath}
\usepackage{color,soul}

\usepackage{multirow}
\usepackage{graphics}

\journal{ICT Express}

\begin{document}

\begin{frontmatter}

\title{Resource Allocation in Multicore Elastic Optical Networks: A Deep Reinforcement Learning Approach}

\author[PUCV]{Juan Pinto-Ríos\corref{mycorrespondingauthor}}
\cortext[mycorrespondingauthor]{Corresponding author}
\ead{juan.pinto.r@pucv.cl}

\author[PUCV]{Felipe Calderón}
\author[PUCV]{Ariel Leiva}
\author[PUCV]{Gabriel Hermosilla}
\author[UCL]{Alejandra Beghelli}
\author[UAI]{Danilo Bórquez-Paredes}
\author[UTFSM]{Astrid Lozada} 
\author[UTFSM]{Nicolás Jara}
\author[UTFSM]{Ricardo Olivares}
\author[UDEC]{Gabriel Saavedra}

\address[PUCV]{School of Electrical Engineering, Pontificia Universidad Católica de Valparaíso, Av. Brasil 2950, Valparaíso 2362804, Chile}

\address[UCL]{Optical Networks Group, Department of Electronic and Electrical Engineering, University College London, WC1E 7JE, UK}

\address[UAI]{Faculty of Engineering and Sciences, Universidad Adolfo Ibáñez, Av. Presidente Errázuriz 3485, Santiago 7941169, Chile}

\address[UTFSM]{Department of Electronic Engineering, Universidad Técnica Federico Santa María, Av. España 1680, Valparaíso 2390123, Chile}

\address[UDEC]{Electrical Engineering Department, Universidad de Concepción, Víctor Lamas 1290, Concepción 4070409, Chile}

\begin{abstract}
A deep reinforcement learning approach is applied, for the first time, to solve the routing, modulation, spectrum and core allocation (RMSCA) problem in dynamic multicore fiber elastic optical networks (MCF-EONs). To do so, a new environment - compatible with \textit{OpenAI's Gym} - was designed and implemented to emulate the operation of MCF-EONs. 
The new environment processes the agent actions (selection of route, core and spectrum slot) by considering the network state and physical-layer-related aspects. The latter includes the available modulation formats and their reach and the inter-core crosstalk (XT), an MCF-related impairment. If the resulting quality of the signal is acceptable, the environment allocates the resources selected by the agent. After processing the agent's action, the environment is configured to give the agent a numerical reward and information about the new network state.
The blocking performance of four different agents was compared through simulation to 3 baseline heuristics used in MCF-EONs. Results obtained for the NSFNet and COST239 network topologies show that the best-performing agent achieves, on average, up to a four-times decrease in blocking probability concerning the best-performing baseline heuristic methods.  
\end{abstract}


\end{frontmatter}



\section{Introduction}\label{sec:introduction}
Due to the ever-increasing traffic demand~\cite{cisco},
new solutions to avoid the eventual capacity exhaustion of current core optical networks have been proposed. They range from improved efficiency in the usage of currently installed resources~\cite{5307468} 
to expand the optical network capacity~\cite{Mizuno:16}. 
One of these solutions is obtained by combining the capacity increase of multicore optical fibers (MCF)~\cite{Saitoh:16} 
with the efficient spectrum usage offered by dynamic elastic optical networks (EONs)~\cite{unknown}. We denote them together as dynamic MCF-EONs.

MCF extend the fiber capacity by adding multiple cores within the same cladding. Thus, the capacity of a single fiber is significantly increased given that each core can be considered as an extra optical medium~\cite{AWAJI2013617}. EONs~\cite{10.1117/1.OE.57.11.110802} divide the spectrum into narrow slots called frequency slot units (FSU), usually of 12.5 GHz width~\cite{Zhou:16}. 
In EON communication, each connection uses as many adjacent slots as needed, thereby improving the spectral usage efficiency~\cite{7288388}. Under dynamic operation, EONs~\cite{7193515} can establish and release connections on-demand. 

One of the main challenges of dynamic MCF-EONs is the design of efficient routing, modulation, spectrum and core assignment (RMSCA) strategies for establishing optical connections with as low blocking probability as possible. Most RMSCA proposals use heuristic approaches that consider the impact of inter-core crosstalk (inter-core XT) on optical signal quality, as described in~\cite{unknown,6958538,6959250,8761380}. Although heuristics are computationally simple, they cannot guarantee optimal solutions~\cite{Zerovnik_2015} 
and their performance depends on the ability of the designer to detect the best set of rules defining the heuristic behaviour.  

In recent years, deep reinforcement learning (DRL) techniques have been applied to solve resource allocation problems in dynamic elastic optical networks to overcome the drawbacks of rule-based systems. For example, DRL was applied to solve the routing, modulation and spectrum assignment (RMSA) problem in single domain EONs ~\cite{Chen:20, Chen_2019,natalino2020optical},  multi-domain EONs~\cite{Li_OFC2020}, multiband EONs~\cite{9492435,9492334} and survivable EONs operating under shared protection~\cite{Luo}; the problem of energy-efficient traffic grooming in fog-cloud EONs~\cite{Zhu_OFC2020} and the problem of establishing and reconfiguring multicast sessions in EONs~\cite{9502141}. 
 Only one previous work has studied the application of DLR on MCF-EONs~\cite{9333250}, but this work focused on fixed-grid networks. In this paper, we extend the work reported in~\cite{Chen:20, Chen_2019,natalino2020optical} by applying DRL for the first time to dynamic MCF-EONs.

A DRL system can be summarized as an agent (an entity equipped with a learning algorithm) that - during its training phase - learns to make good decisions by interacting with an environment~\cite{mnih2015humanlevel,8585411}. In the context of RMSCA, the agent must learn to allocate optical resources to connection requests such that they are not blocked due to physical impairments or lack of spectral continuity or contiguity in the chosen route. A reasonable allocation decision makes the environment give the agent a high-value reward. In resource allocation in optical networks, the environment is programmed to represent the state, operation and constraints of a dynamic optical network. When a connection request arrives during the training phase, the agent decides what resources to allocate. At the beginning of its training, the agent makes arbitrary decisions (exploration process). Then, the environment determines whether the set of resources identified by the agent is feasible and gives the agent feedback about the quality of its decision. This information, stored in the experience buffer of the agent, allows the agent to learn. As a result, it starts to select better actions (exploitation process) for future requests. Better actions result in the agent earning a high cumulative reward. After an agent has finished its training stage, it can be evaluated (testing stage) by having it process connection requests it has never received before.

In the context of dynamic MCF-EONs, with the exception of~\cite{9333250}, only supervised machine-learning techniques have been applied so far. These consist of techniques for making inferences based on expert-labelled data. Thus, instead of taking actions, supervised learning algorithms perform estimations or classifications ~\cite{10.1007/978-3-319-56991-8_32}. 

For example, the authors of~\cite{XIONG201999, xiong2020deep} used supervised learning to predict future connection requests in dynamic MCF-EONs to perform a crosstalk-aware resource allocation in advance. Instead, the authors in ~\cite{8307057} used machine learning to estimate the inter-core XT to then execute a crosstalk-aware allocation algorithm. All these studies have used machine learning as an auxiliary process to improve the assignment, either by predicting future traffic or transmission quality. In none of them, machine learning had direct participation in the decision-making related to resource allocation.

To the best of our knowledge, there are no previous studies applying DRL to solve the RMSCA problem in dynamic MCF-EONs. In this paper, we present, for the first time, the implementation and testing of a new dynamic MCF-EON environment where four different DRL agents are trained to solve the RMSCA problem. The results obtained by the best performing agent are then compared to 3 baseline heuristics.

The rest of this article is organized as follows: Section II presents the DRL system developed, Section III describes the performance evaluation experiments, and Section IV concludes the paper.

\section{DRL for dynamic MCF-EONs}

The implementation process of any DRL system can be done in two stages:
\begin{itemize}
\item Stage 1: Environment Design and Implementation. The environment is a program that receives the agent´s action, processes it, and sends back feedback. The specific feedback depends on the results of the agent´s action on the environment. The environment must consider the characteristics and constraints of the existing system to process the action. In the case of an optical network, the environment must manage information about the network topology and status and model the network operation (including physical phenomena related to the signal transmission and spectrum allocation constraints).

\item Stage 2: Agent Training. The agent must first acquire knowledge about the environment. This training is done by exploration and exploitation. When exploring, the agent selects random actions to learn how the environment reacts and stores such knowledge. When exploiting stored knowledge, the agent makes informed decisions to select the following action. During exploration and exploitation, the agent receives feedback from the environment, which the agent uses to update its knowledge. In this way, the agent´s training progresses.
\end{itemize}

In the following, these two stages are described in detail in the context of dynamic MCF-EONs.

\subsection{Stage 1: Environment design and implementation}

In this work, the toolkit \textit{Optical RL-Gym}, developed by Natalino and Monti~\cite{natalino2020optical} to facilitate the implementation and 
replicability of deep reinforcement learning environments for optical networks, was extended by creating a new environment: \texttt{DeepRMSCAEnv}. Such an environment encapsulates all the necessary functions to simulate an MCF-EON. 

\begin{figure*}[h!]
    \centering
    \includegraphics[scale=.55]{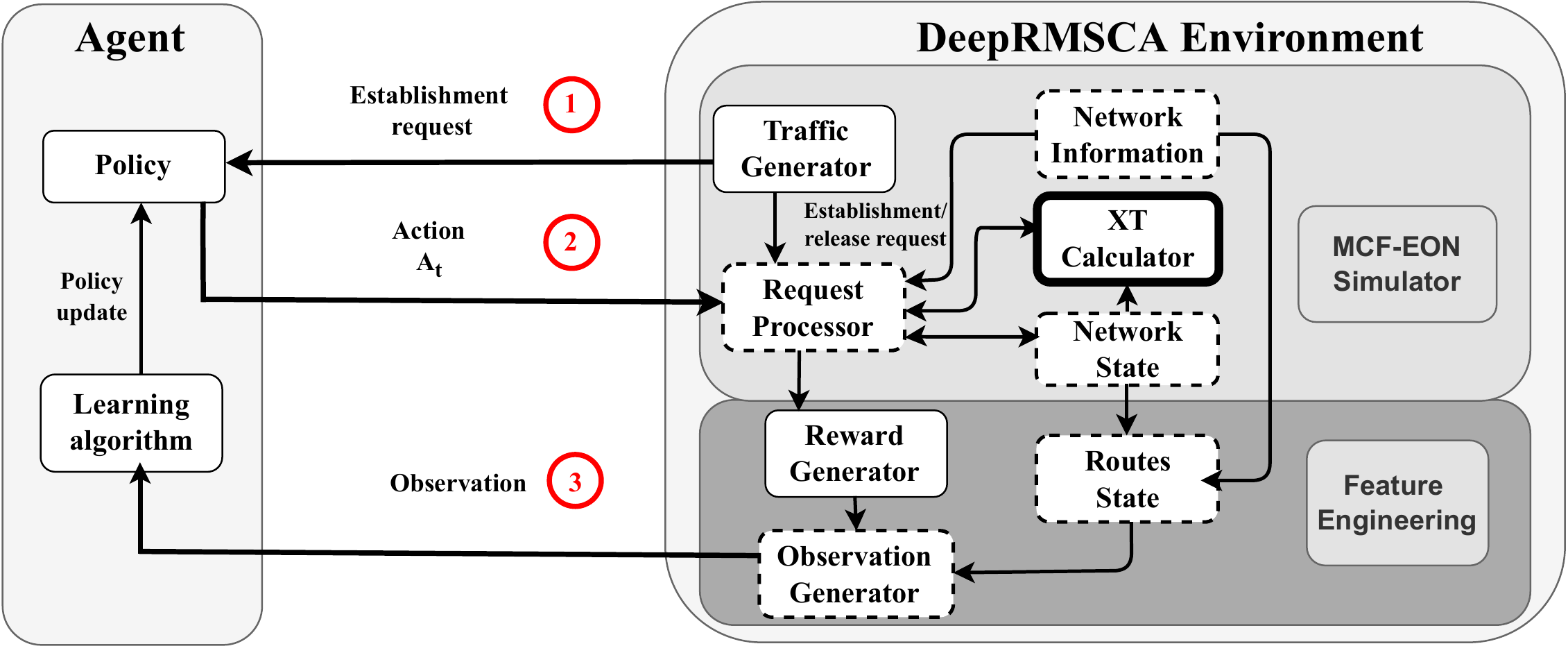}
    \caption{Interaction between an DRL agent and the MCF-EON environment developed: DeepRMSCAEnv.} 
    \label{fig:fig1}
\end{figure*}

The right part of Figure~\ref{fig:fig1} shows a schematic of the implemented environment, including its main components and interactions. Dashed and thick lines modules are modules from the Optical RL-Gym toolkit that had to be modified and developed from scratch, respectively, to model an MCF-EON environment correctly. 

The environment can be considered an event-driven dynamic MCF-EON simulator and a feature engineering module. The former is responsible for processing the connection requests according to the agent´s action and sending the relevant information to the feature engineering module. The latter is responsible for preparing and sending feedback to the agent (reward and observation). 

The dynamic MCF-EON simulator consists of five components. Two of these store data about the network: 
\begin{itemize}
    \item Network information. This component stores the graph representation of the network and the link capacity, considering the multicore nature of links. It also stores the \texttt{K} alternatives routes for each source-destination pair, the modulation format used as a function of the route length distance 
    and the network information, coded as in~\cite{Chen_2019}.  
    \item Network state. This component stores the state of each slot (available or used) for each network link and core. 
\end{itemize}

The remaining 3 components perform specific tasks:

\begin{itemize}
    \item Traffic generator. This component is responsible for the random generation of connection establishment and release requests. Connection establishment requests, defined by the triplet $(src, dst, b)$, are sent to the agent and the request processor component, where $src$ is the source node, $dst$ is the source node the destination node and $b$ is the bitrate. The connection release requests are sent only to the request processor.
    \item Request processor. This component receives several inputs. The first two are the connection establishment or release request and the agent´s action (in the case of an establishment request). When a connection establishment request is received, the request processor waits for the agent´s action (route, core, initial slot). Once the agent´s action is received, the Request Processor first determines the number of slots the connection requires. To do so, the most efficient modulation format that ensures a QoT~\cite{5465362} is first selected (QoT has been transformed into a maximum reach, as shown in Table 1). The calculation of the number of slots is the same described in Section II of~\cite{Chen_2019}. 
    \begin{table}[h]
    \caption{Maximum reach for each modulation format~\cite{7302513}}
    \begin{center}
     \begin{tabular}{||c c||} 
     \hline
     Modulation format & Max. reach (km)\\ [0.45ex] 
     \hline\hline
     64QAM & 250 \\ 
     \hline
     32QAM & 500 \\
     \hline
     16QAM & 1000 \\
     \hline
     8QAM & 2000 \\
     \hline
     QPSK & 4000 \\
     \hline
     BPSK & 8000 \\
     \hline
    \end{tabular}
    \label{Umbral}
    \end{center}
    \end{table}
    
    Next, it checks the network topology (input from the Network information module) and the network state (input from the Network state module) to evaluate the availability of the resources selected by the agent. It also obtains information from the XT calculator component regarding the feasibility of the allocation in terms of crosstalk. If resources are available and a positive answer is received from the XT calculator, then resources are allocated, and the corresponding information is updated on the network state module. Information about a successful establishment is also sent to the reward generator. 
    If resources cannot be allocated, information about the failed establishment is sent to the reward generator component only. When a connection release is received, the request processor component updates the network state module to make the released resources available.
    \item XT calculator. This component calculates the inter-core crosstalk (XT), defined as the interference between optical connections in neighbouring cores using the same frequency slots. It receives information about the resources selected by the agent (length of the links composing the route and core) and evaluates the XT. For generic MCF systems, with any number of cores in any geometric arrangement, the steps to calculate the mean XT affecting a connection established in core $i$ are as follows:
    \begin{itemize}
        \item Calculate the mean XT per unit of length between core $i$ and adjacent core $j$, $h_{i,j}$ as :
    
     \begin{equation}
     \label{eq:h}
     h_{i,j}=\frac{2k^2 r}{\beta \Lambda_{i,j}}
     \end{equation}
    where $k$, $r$, $\beta$ and $\Lambda$ are the coupling coefficient, radius of curvature (or bending), constant propagation and the distance between cores $i$ and $j$ respectively.
    
    \item Calculate the total mean XT affecting core $i$, $XT_{i}$, by adding the crosstalk contribution of all its adjacent cores. That is:
     \begin{equation}
     \label{eq:XTi}
     XT_{i}=\sum_{j=1}^{n} h_{i,j}\cdot L
     \end{equation}
    where $n$ is the number of cores adjacent to core $i$ and $L$ the length of the link.
    \end{itemize}
    For the specific case where cores follow a triangular or hexagonal geometric arrangement and different pairs of cores are equidistant, equation~\eqref{eq:XTHex} has been found to be a better approximation to calculate $XT_{i}$~\cite{unknown}:
 
    \begin{equation}
    \label{eq:XTHex}
     XT_{i}=\frac{n-n\cdot exp[-(n+1)\cdot hL]}{1+n\cdot exp[-(n+1)\cdot hL]}
    \end{equation}

 where, as in equation~\eqref{eq:XTi}, $n$ represents the number of cores neighbouring $i$, and $L$ is the length of the link. The term $h$ is given by equation~\eqref{eq:h} (sub-indices have been dropped since the distance between all core pairs is assumed to be the same).
    
   An XT threshold value for different modulation formats is defined in~\cite{7302513}~\cite{ZHAO2017249} such that the signal quality is acceptable. If  XT exceeds this predefined threshold (summarized in Table~\ref{Umbral}), a negative answer is sent to the request processor (-1). Otherwise, a positive answer is sent (1). 
\end{itemize}

\begin{table}[h]
\caption{XT threshold for each modulation format~\cite{ZHAO2017249}}
\begin{center}
 \begin{tabular}{||c c||} 
 \hline
 Modulation format & XT threshold (dB)\\ [0.45ex] 
 \hline\hline
 64QAM & -34 \\ 
 \hline
 32QAM & -27\\
 \hline
 16QAM & -25 \\
 \hline
 8QAM & -21 \\
 \hline
 QPSK & -18 \\
 \hline
 BPSK & -14 \\
 \hline
\end{tabular}
\label{Umbral1}
\end{center}
\end{table}

The feature engineering module in Figure~\ref{fig:fig1} is responsible for preparing the information to be sent back to the agent. It is made of three components:

\noindent\textbf{Reward generator.} This component calculates the numerical reward to be sent to the agent depending on the information received from the request processor component. In this work, a successful resource allocation returns a reward equal to 1 and a failed allocation equal to -1. Connections can be rejected due to lack of spectrum resources along the path selected by the agent, because of crosstalk among cores exceeding the predefined threshold or because the length of the path selected by the agent is longer than the maximum optical reach of any modulation format (such limit depends on the modulation format and a bit-error-rate threshold, as in Section III, Table 3 of~\cite{5534599}). 

\noindent\textbf{Routes state.} This component receives the routing information from the network information component and the utilization state of the slots in such routes from the Network State component. This information is then consolidated in a 1D vector made of \((K\cdot C\cdot J)\) elements, where \(K\) is the number of alternative routes, \(C\) is the number of cores and \(J\) is the number of blocks with enough available slots to establish the connection request being processed. A block is a set of contiguous available slots. 

\noindent\textbf{Observation generator.} This component consolidates - for the core selected by the agent - the following information in a 1D vector, taking as a basis the observation vector used in ~\cite{Chen_2019}: source node (one-hot encode), destination node (one-hot encode), holding time, reward, the number of slots requested, routes state, the action taken, and the reward for that action.

\subsection{Stage 2: Agent training}
The left side of Figure~\ref{fig:fig1} shows the interaction between the agent and the \texttt{DeepRMSCAEnv} environment during the training stage. 

\begin{figure*}[h!]
    \centering
    \includegraphics[scale=.55]{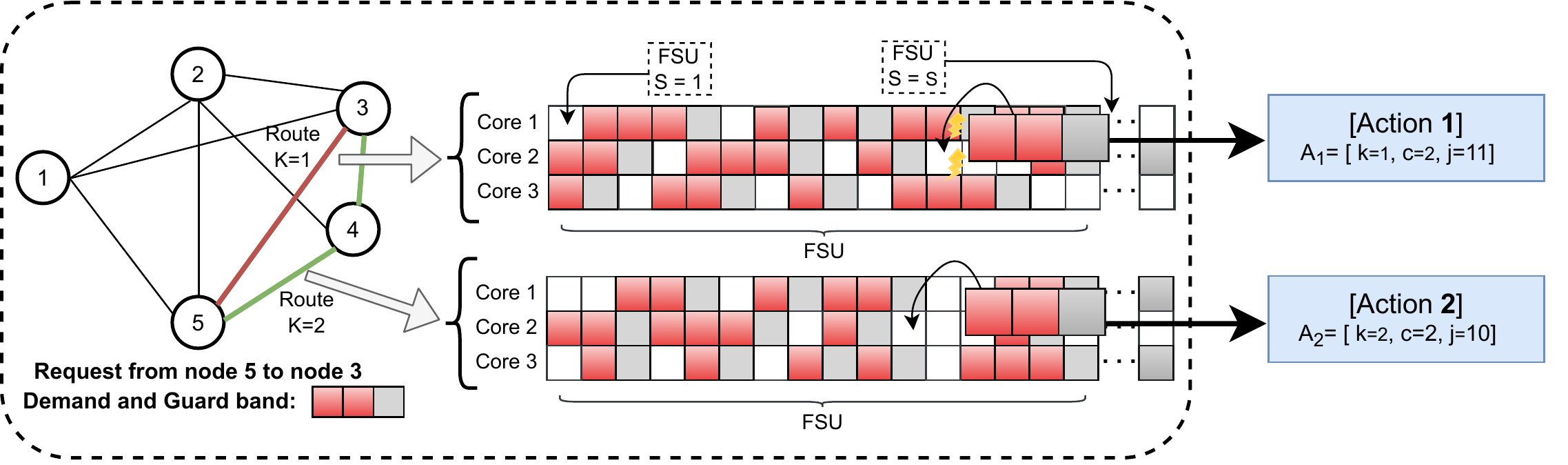}
    \caption{Example of connection request from node 5 to 3, requesting 3 slots (2 for data, 1 as guard band). K=2, meaning 2 routes. The routes spectral use is represented by white blocks (available FSUs) and red and gray ones (occupied for data and as guard bands, respectively).} 
    \label{fig:figXAI1}
\end{figure*}

The agent aims to maximize its long-term reward. That is, selecting actions leading to the highest number of connection requests established. To achieve this goal, the agent is built considering two main components:

\noindent\textbf{Policy.} This component is where the behaviour of the agent is embedded. At a given time $t$, receives a connection establishment request as input along with a summary of the state of the network~\cite{Chen_2019} and an action $A_t$ is outputted. The action is defined by 3 integer numbers: a path identifier $k$ (selected out of \textit{K} possible pre-computed routes), a core identifier \textit{c} (selected out of \textit{C} possible cores) and the identifier of the initial frequency slot \textit{j} of the block selected. These values define which route, core and spectrum resources should be assigned to each request. As the agent successfully allocates more connection requests, the policy becomes better. At the end of the training, the policy is expected to allow the agent to define which action has the highest probability of not being blocked. 
Figure~\ref{fig:figXAI1} shows a simplified example of two possible actions that might be taken by the agent, given a specific network state.

On the left part of the figure, 5-node network topology and a connection establishment request of 2 slots between nodes 5 and 3 are shown. The demand is represented by the red boxes (2 slots in this case) plus the grey box (1 slot used as a guard band). The amount of slots required to serve the connection (red squares) is determined by the modulation format, using the same method presented in~\cite{Chen_2019}. One guard band of 1 slot is considered for each connection request to achieve a good trade-off between the quality of transmission and the blocking probability~\cite{8024956}. 

Let us assume the network is equipped with three cores per link, and the agent can select either route 1 ($k=1$), represented by the red link in the topology, or route 2 ($k=2$) by the green links. In addition to a route, the agent must also select a core and a slot. On the right side of the figure, the spectrum utilization of both routes is shown. Red and grey squares represent used FSUs. A row of squares represents the slot utilization in a specific core for a specific route. Thus, the three rows on the upper and lower part of the figure represent the slot utilization on the three cores of the first and second routes, respectively.  
    
If the agent selects Action 1, depicted in the upper part of the figure, then action $A_t=[1,2,11]$ is sent back to the environment, signalling that the agent selects slot 11 as the initial slot on route 1 in core 2 to establish the connection. The thunderbolt symbol in route 1 represents the presence of crosstalk exceeding the acceptable threshold. In this case, the request will be rejected, and a reward of -1 will be sent to the agent.
Instead, if the agent selects Action 2, depicted on the lower part of the figure, then action  $A_t=[2,2,10]$ is sent back to the environment. This action leads to a successful connection establishment, and the agent receives a reward equal to 1. During the training stage, the policy component should be updated to select Action 2 over Action 1 (in this case), leading to a higher reward.
    
\noindent\textbf{Learning algorithm.} This component receives the rewards and observations from the environment and, based on that information, updates the policy to produce actions that maximize the expected cumulative long-term reward. In this study, we consider learning algorithms compatible with the action space considered. The action space used has a multi-discrete nature because the action is defined by multiple discrete values (route, core and slot identifier). Thus, the learning algorithms available in the  \texttt{Stable-Baselines}~\cite{stable-baselines} library that were compatible with a multi-discrete space state were selected (as also done in~\cite{Chen_2019, natalino2020optical}). These are:
   
    \begin{itemize}
     
    \item Advantage Actor-Critic (A2C)~\cite{mnih2016asynchronous} and Actor-Critic using Kronecker-Factored Trust Region (ACKTR)~\cite{wu2017scalable}: These are approaches based on the actor-critic algorithm~\cite{mnih2016asynchronous}, which has two interacting neural networks. The actor uses a dense neural network to process and update the policy obtained. The critic uses a separated neural network to evaluate the quality of the policy by calculating the 'value function'~\cite{mnih2015humanlevel}. Both algorithms differ in how they update their neural networks' weights. A2C does that by using the feedback the critic's network gives to the actor's network, whilst ACKTR uses a Kronecker-factored approximation~\cite{wu2017scalable}, which is a method that optimizes the stochastic gradient descent.
    \item Proximal Policy Optimization(PPO2)~\cite{schulman2017proximal} and Trust Region Policy Optimization (TRPO)~\cite{pmlr-v37-schulman15}: These learning algorithms use only one neural network, whose weights are updated based on the policy gradient descent. They differ in the way the policy gradient descent is approached.  
    TRPO avoids sudden changes in the neural network weights, updating only those that do not differ by a greater distance than what the Kullback–Leibler restriction (relative entropy)~\cite{pmlr-v37-schulman15} allows. Instead, PPO2 does not impose limits on the neural network weights' changes to optimise the policy's descent curve. 
    
    \end{itemize}

\section{Performance Evaluation}

 Table~\ref{tab:parameters_sim} lists the values of the main parameters used to train the agents. In terms of network parameters, we consider two topologies: the NSFNet and the COST239. For each one, we assume 100 FSUs and 3 cores arranged in a triangular geometry   
 per link, and the available modulation formats are BPSK, QPSK, 8-QAM and 16-QAM. These simplifications have been considered due to memory constraints. The same number of slots was considered in \cite{9492435}. As in ~\cite{8717574}, we use ~\eqref{eq:XTHex} to calculate the XT.
 Regarding the traffic characteristics, we assume a fully dynamic behaviour, where connection establishment requests arrive as a Poisson process and connection holding times follow a negative exponential distribution. The bitrate associated to each connection is uniformly selected from the range [25-100] Gbps, as in \cite{Chen_2019}.  
 Finally, regarding the agents (one per learning algorithm), they will select one out of 5 pre-computed routes, one out of 3 cores and the number of FSU needed for the connection considering a total of 100 FSUs. Agents will be trained in episodes made of 50 connection requests each (to simplify backpropagation in the dense neural network used by the agent by delivering small batches of data continuously), and the whole training session will consider a total of 160,000 connection requests. The parameters of the four agents will be the ones set by default in the agent's library \textit{Stable Baselines}\cite{stable-baselines} 
 The DRL system developed is available in a Git repository~\footnote{The new environment, under the name \texttt{DeepRMSCAEnv}, is available at: \url{https://gitlab.com/IRO-Team/deeprmsca-a-mcf-eon-enviroment-for-optical-rl-gym/}}. 
 

\begin{table}[h]
\centering
\caption{Network, traffic and training parameters} \label{tab:parameters_sim}
\resizebox{.48\textwidth}{!}{
\begin{tabular}{| l | l |}
\hline
\textit{Parameters} & \textit{Value} \\
\hline \hline
\textbf{Network Parameters} & \\
Topologies & NSFNet~\cite{10.1145/55482.55502} and COST239~\cite{11.1145/55482.55502}\\
Number of cores & $3$\\
Number of FSU by link  & $100$ \\ 
Modulation Formats & BPSK, QPSK, 8-QAM, 16-QAM\\
\hline
\textbf{Traffic Parameters} & \\
Bit rates [Gb/s] & Uniformly distributed in [25-100] Gbps \\ 
\hline
\textbf{Agent Training Parameters} & \\
Pre-computed candidate routes & $5$ \\
Number of connection requests per episode & $50$~\cite{Chen_2019}\\
Simulated requests per training & 160,000\\
Agent's learning algorithm parameters & By-default~\cite{stable-baselines}\\
\hline
\end{tabular}
}
\end{table}
\subsection{Preliminary Training Results}
The agents TRPO, PPO2, A2C and ACKTR, were trained with a traffic load of 250 Erlang, as in~\cite{Chen_2019}.

\begin{figure}[h!]
    \centering
    \includegraphics[width=0.49\textwidth]{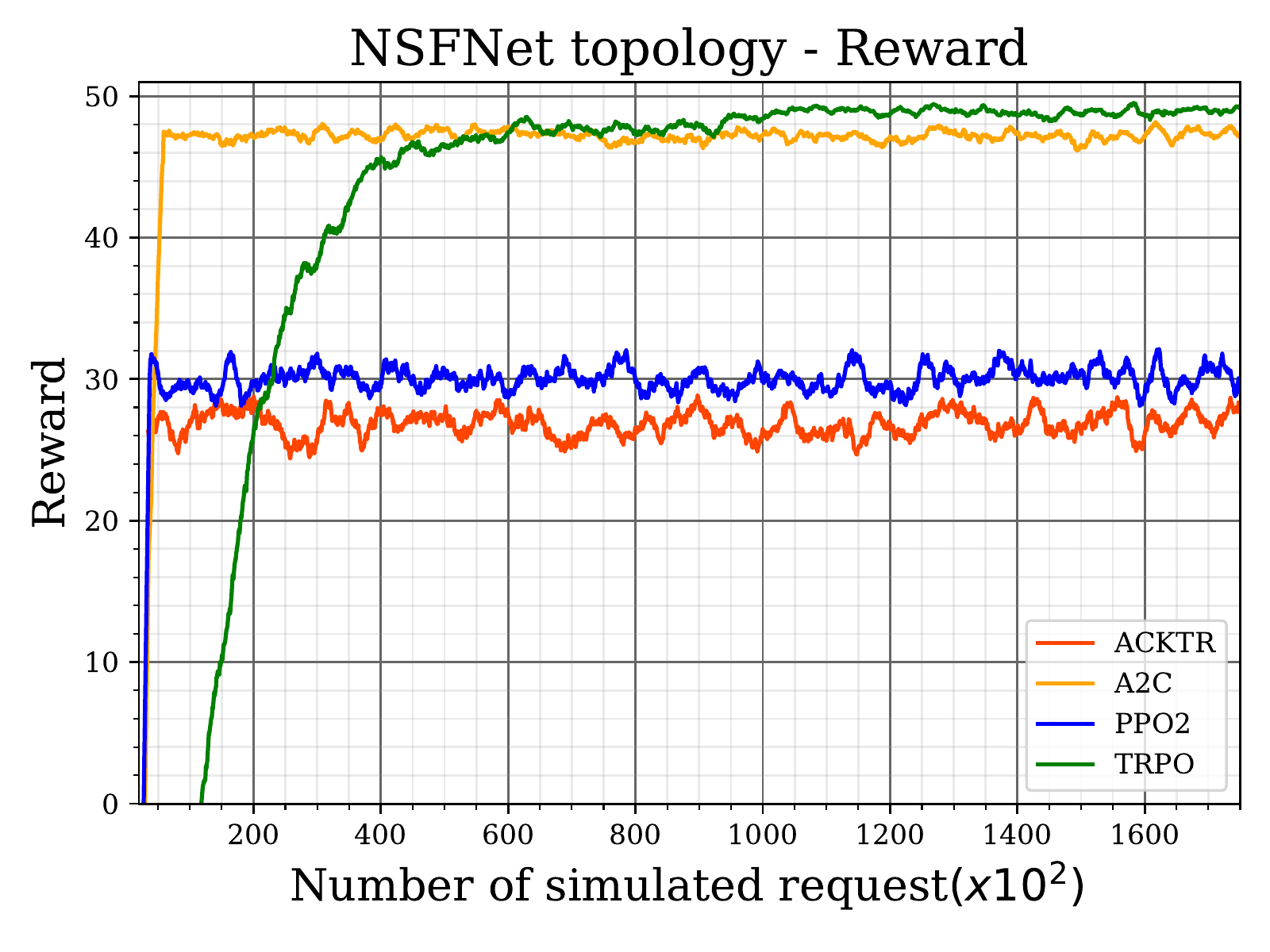}
    \caption{Accumulated reward for the A2C, PPO2, TRPO and ACKTR agents in the NSFNet topology}
    \label{fig:fig4}
\end{figure}

\begin{figure}[h!]
    \centering
    \includegraphics[width=0.49\textwidth]{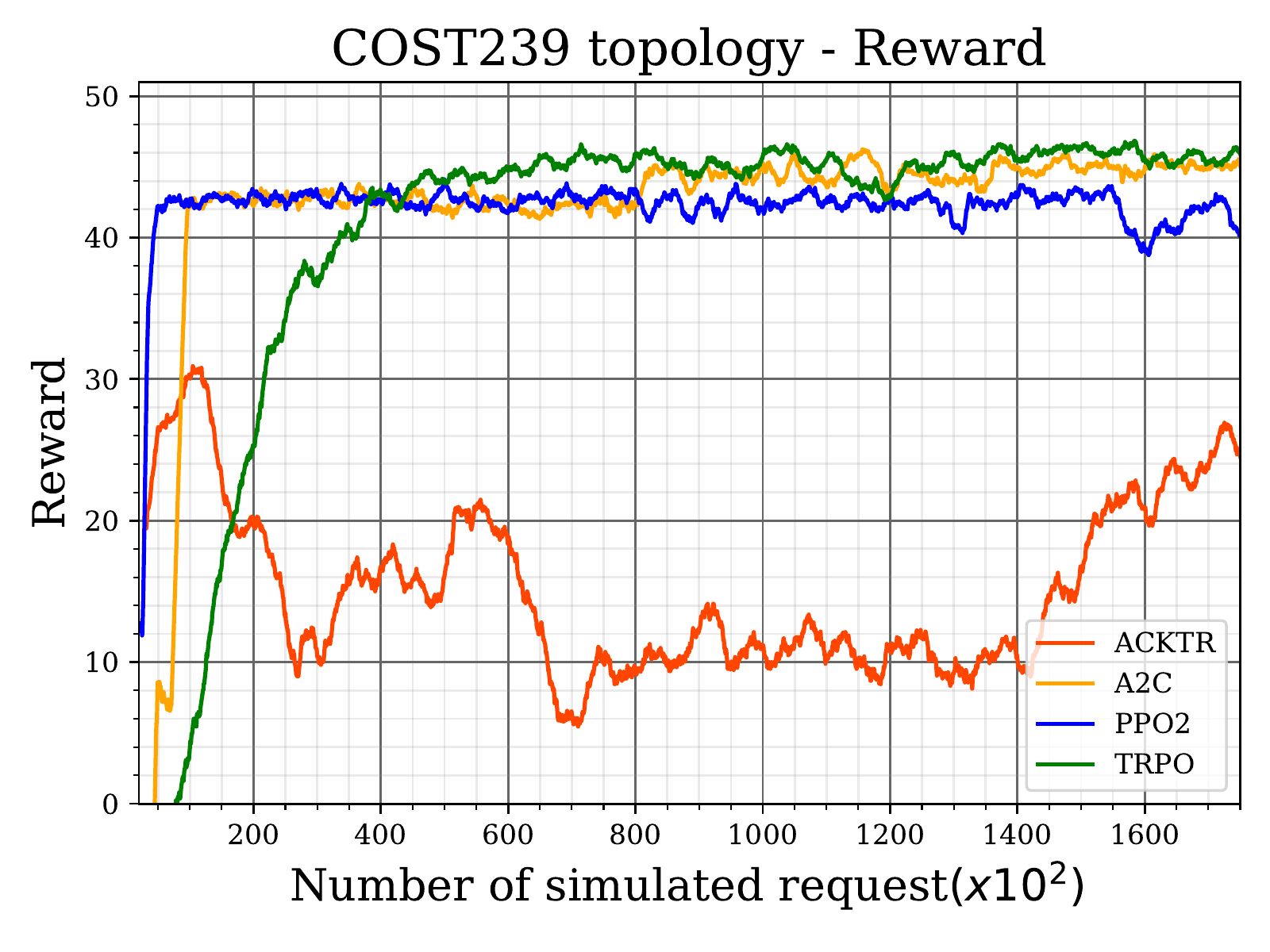}
    \caption{Accumulated reward for the A2C, PPO2, TRPO and ACKTR agents in the COST239 topology }
    \label{fig:fig5}
\end{figure}

Figures \ref{fig:fig4} and \ref{fig:fig5} show the reward accumulated by the different agents during their training in the NSFNet and COST239 topologies, respectively. Given that each episode is made of 50 connection requests, the maximum reward achievable by an agent is 50. It can be seen that the A2C and TRPO agents are the only ones reaching values close to the maximum expected reward in both topologies with an average reward of 49 and 47, respectively, with TRPO exhibiting slightly better performance.


\begin{figure}[h!]
    \centering
    \includegraphics[width=0.49\textwidth]{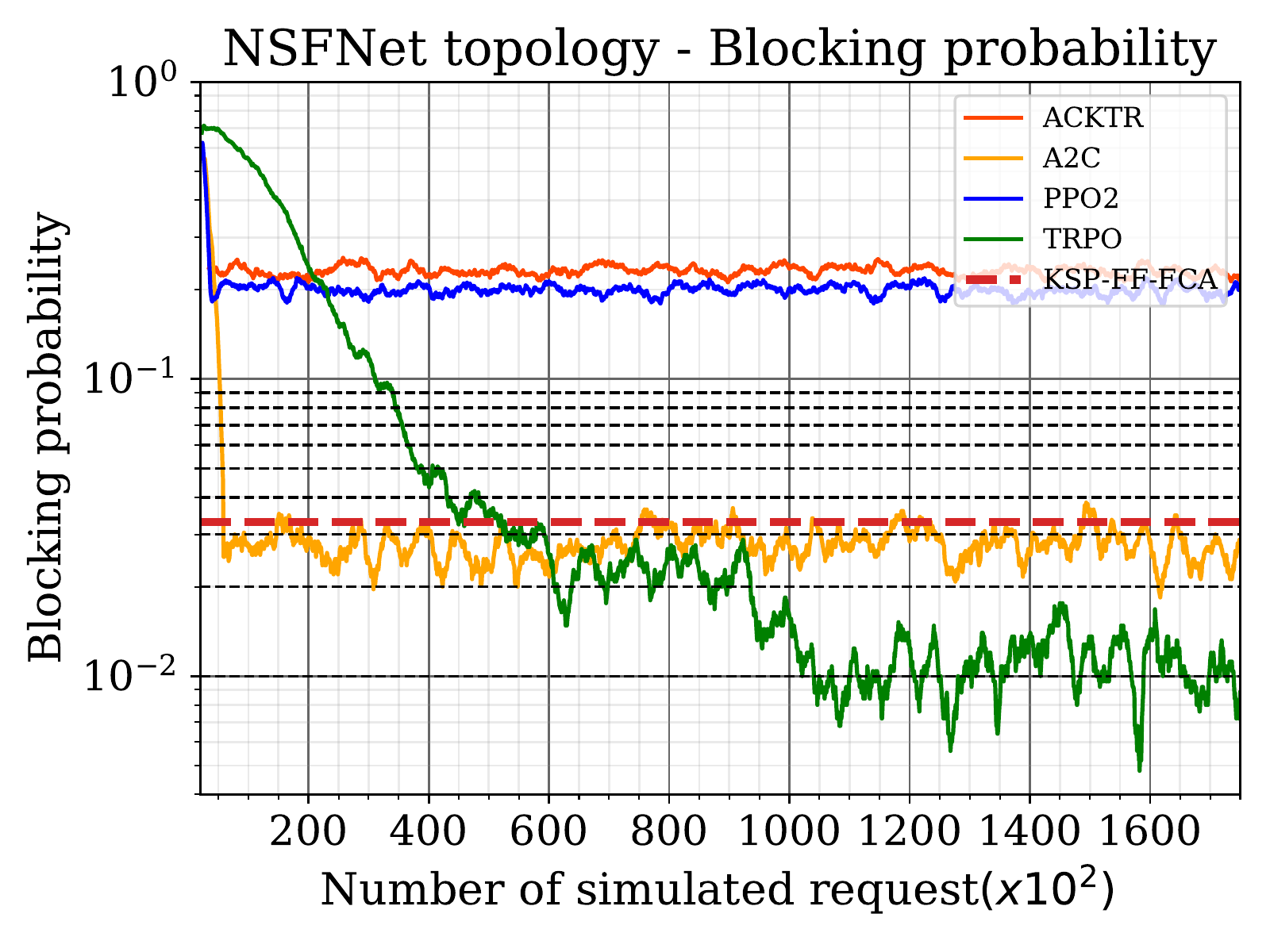}
    \caption{Blocking probability for A2C, PPO2, TRPO and ACKTR in NSFNet Topology }
    \label{fig:fig21}
\end{figure}

\begin{figure}[h!]
    \centering
    \includegraphics[width=0.49\textwidth]{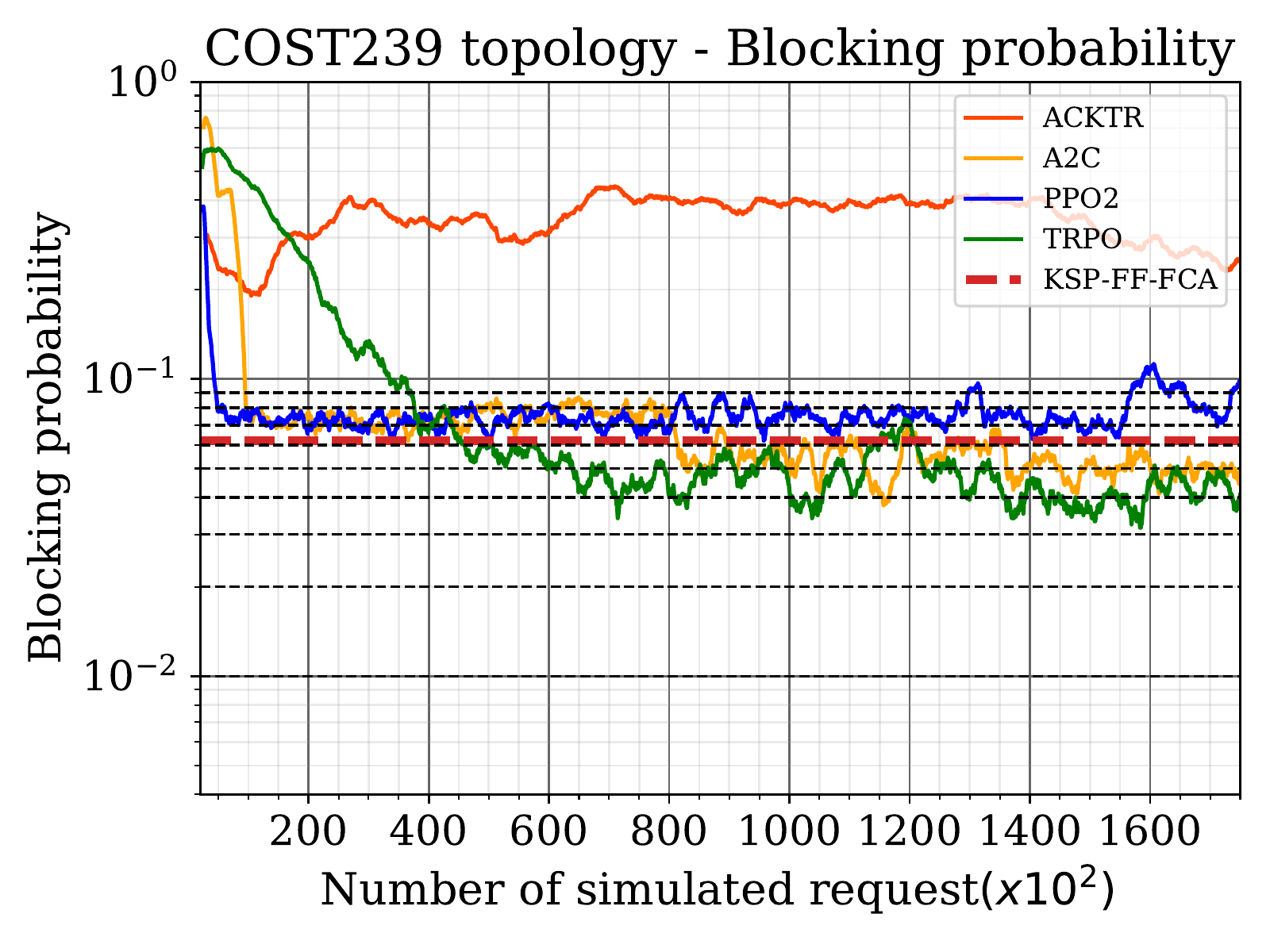}
    \caption{Blocking probability for A2C, PPO2, TRPO and ACKTR in COST239 Topology }
    \label{fig:fig3}
\end{figure}

Figures \ref{fig:fig21} and \ref{fig:fig3} show the evolution of the blocking probability during the training process of the same agents for the NFSNet and COST239 topologies, respectively. For comparison, the dashed red line shows the blocking probability obtained by one of the baseline heuristics, kSP-FF-FCA. This heuristic has a list of 5 pre-computed paths, sorted from shortest (k=1) to longest (k=5). When a connection request arrives, the heuristic attempts to establish the connection in the shortest path of the list (k=1), applying the first-fit policy for spectrum allocation and first fit crosstalk-aware for core allocation, as described in ~\cite{7214205}. The same procedure is repeated for the following path in the list if unsuccessful. After attempting all paths, the connection is rejected if there are no available resources. 

From the figure, we can see that, once the agents are in steady-state, TRPO and A2C agents outperform the heuristic, improving blocking of 24.3\% and 73.9\% for the NSFNet topology and  14.51\% and 38.71\% for the COST239 topology, respectively.  

Given the excellent performance of the TRPO agent in both topologies, in the following section, this agent will be trained for different traffic loads, and then its performance will be contrasted with that of the heuristics selected in ~\cite{BRASILEIRO2020100584}. 



\subsection{TRPO TRAINING RESULTS}

\begin{figure}[h!]
    \centering
    \includegraphics[width=0.49\textwidth]{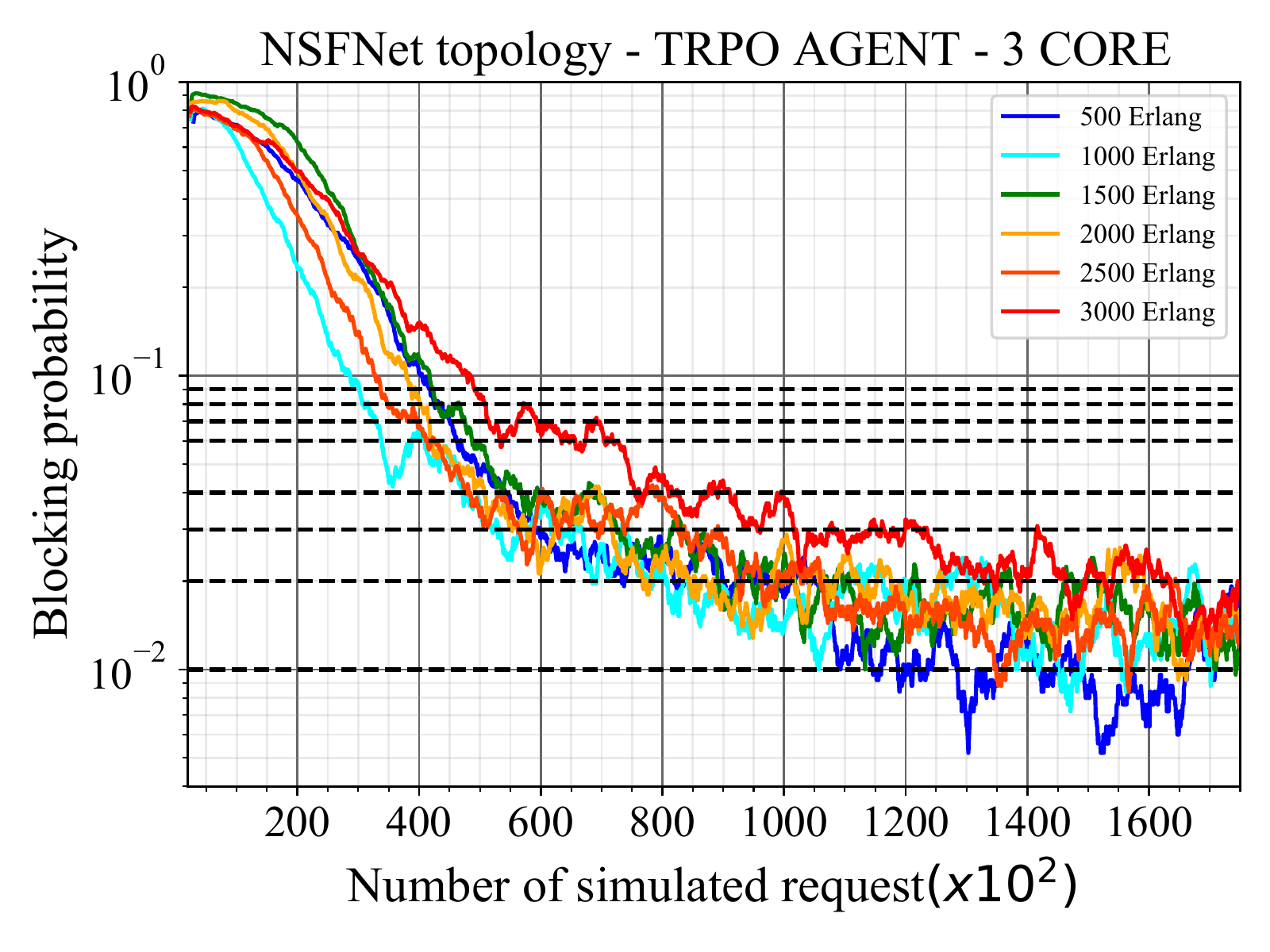}
    \caption{Blocking probability progress for TRPO agent training in NSFNet Topology.}
    \label{fig:fig6}
\end{figure}

\begin{figure}[h!]
    \centering
    \includegraphics[width=0.49\textwidth]{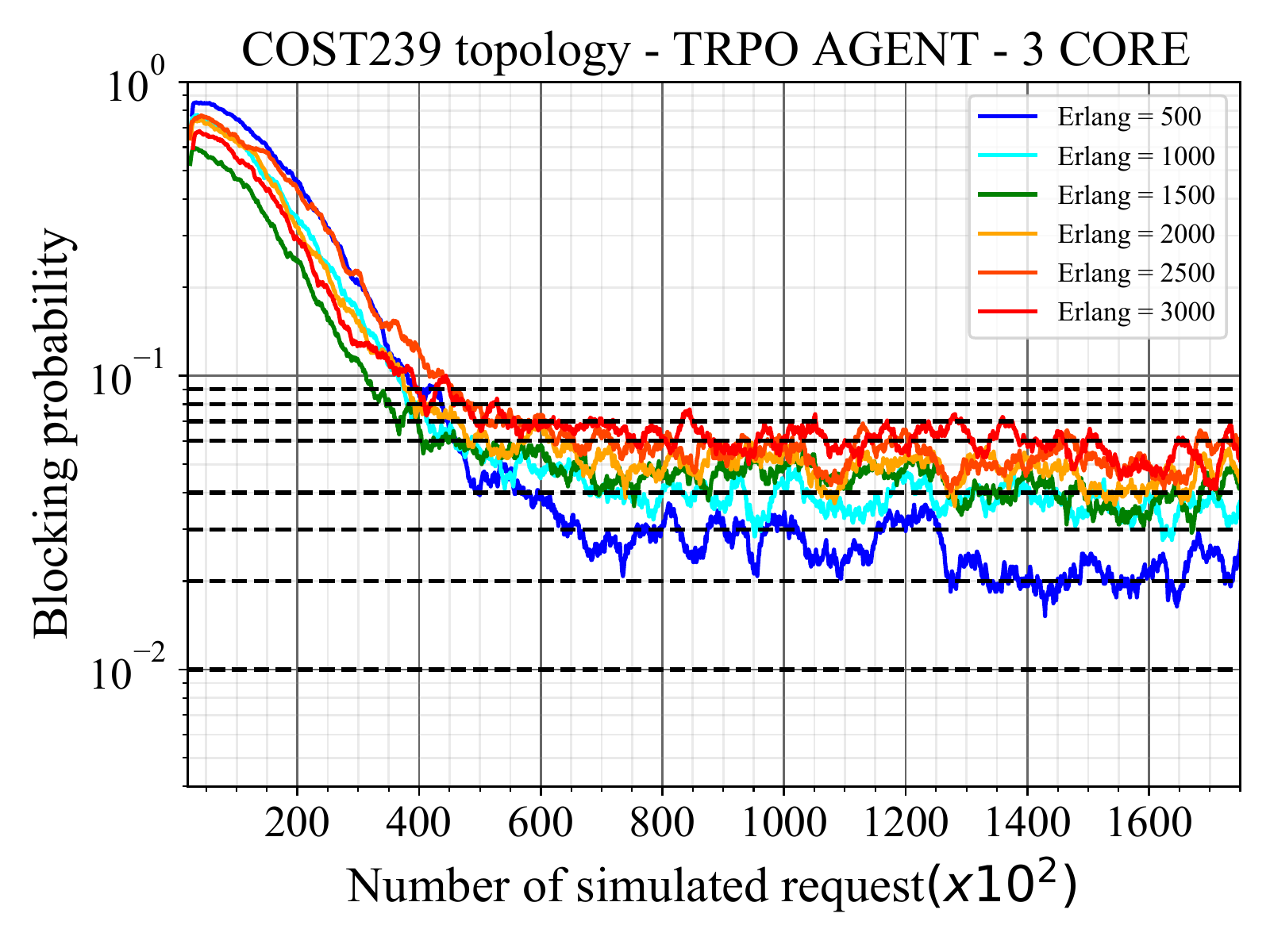}
    \caption{Blocking probability progress for TRPO agent training in COST239 Topology.}
    \label{fig:fig7}
\end{figure}

The TRPO agent was trained for traffic loads between 500 and 3000 Erlang; in steps of 500. Figures~\ref{fig:fig6} and~\ref{fig:fig7} show the evolution of the blocking probability achieved by the TRPO agent as a function of the number of connection requests for different traffic loads for the NSFNet and COST239 topologies, respectively.
It can be seen that the agent exhibits consistent behaviour, with the blocking probability increasing with the traffic load. 


\subsection{TRPO AGENT VS. HEURISTIC: BLOCKING PERFORMANCE}
\begin{figure}[h!]
    \centering
    \includegraphics[width=0.49\textwidth]{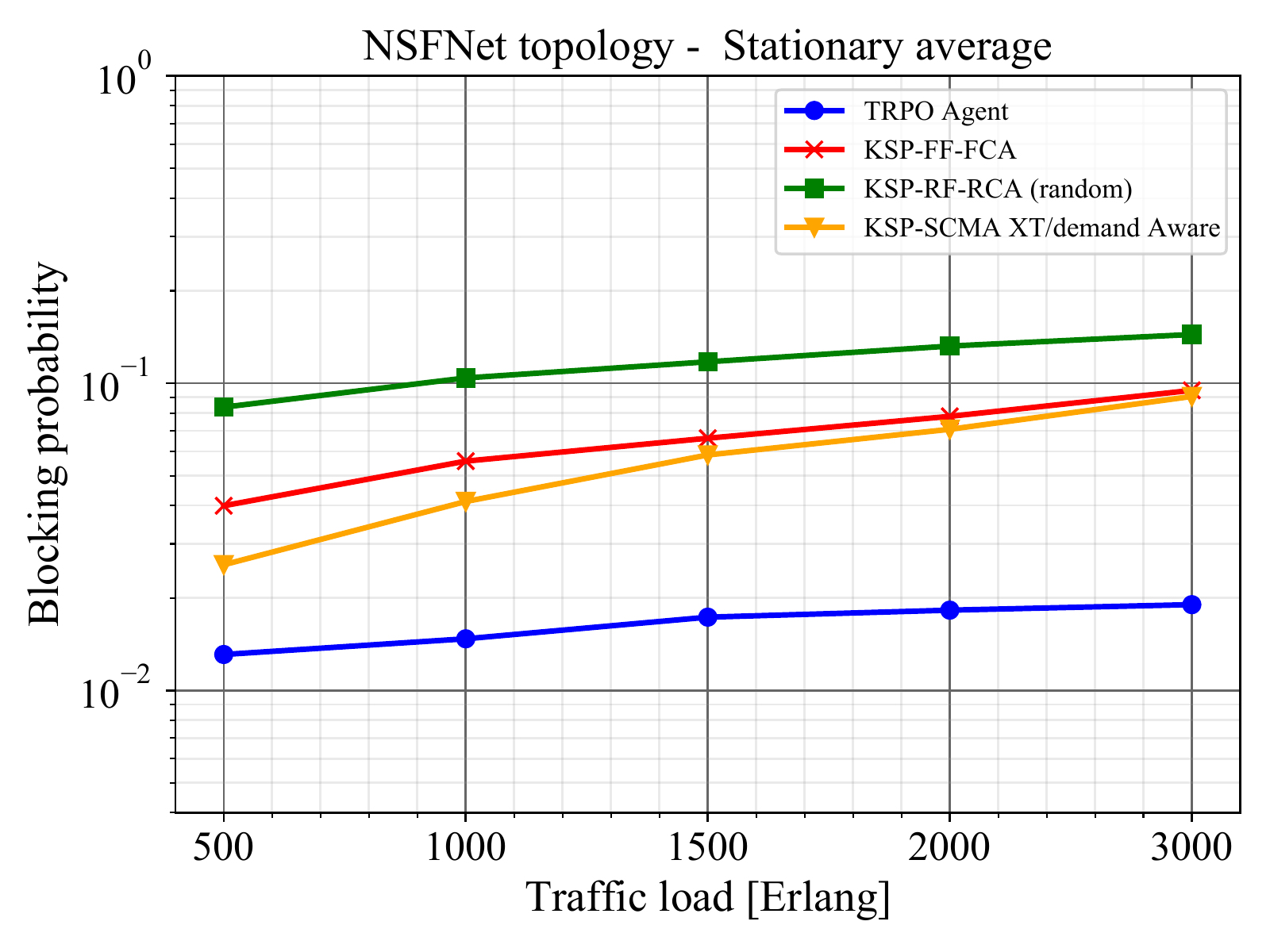}
    \caption{Blocking probability steady average of TRPO agent trained in NSFNet topology.}
    \label{fig:fig9}
\end{figure}

\begin{figure}[h!]
    \centering
    \includegraphics[width=0.49\textwidth]{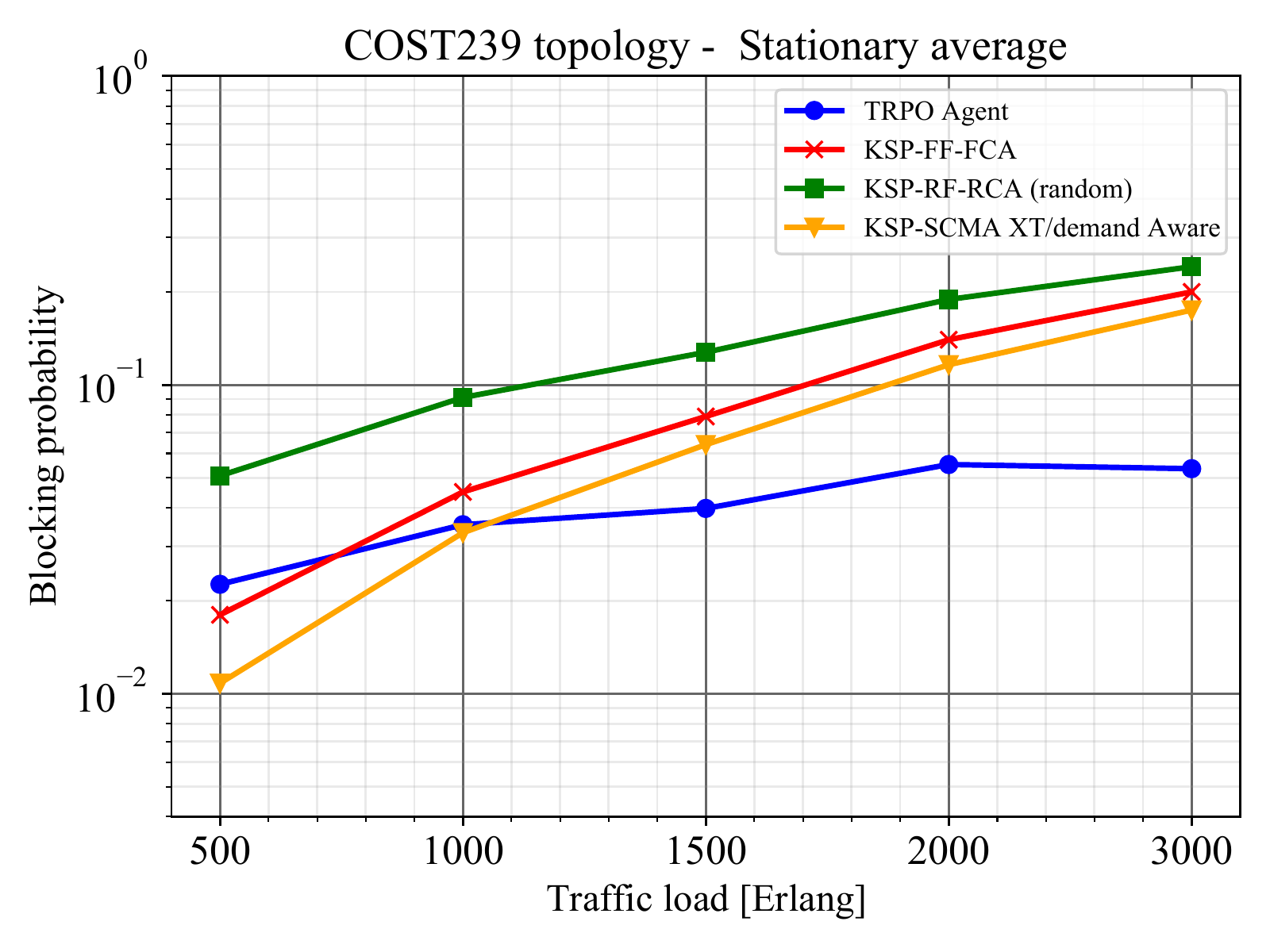}
    \caption{Blocking probability steady average of TRPO agent trained in COST239 topology.}
    \label{fig:fig10}
\end{figure}

Figures~\ref{fig:fig9} and ~\ref{fig:fig10} show the blocking probability achieved by the trained TRPO agent and the same heuristics selected for blocking evaluation in the survey~\cite{BRASILEIRO2020100584}: KSP-FF-FCA, KSP-RF-RCA and KSP-SCMA XT/demand Aware~\cite{6958538} for the NSFNet and COST39 topologies, respectively. The three heuristics apply alternated routing. KSP-FF-FCA uses the First Fit policy to select core and spectrum, KSP-RF-RCA applies a random policy to select core and spectrum, and KSP-SCMA XT/demand aware allocates different parts of the spectrum depending on the Bitrate of the connection request.

Compared to the best performing heuristic, KSP-SCMA XT/demand-aware, a significant improvement in the blocking performance of the DRL approaches is observed. For example, in the NFSNet topology, at the highest load studied, the TRPO agent exhibits a blocking probability of about $1.9\cdot10^{-2}$, about four times slower than the blocking of $8.5\cdot10^{-2}$ achieved by the heuristic. On average, considering both topologies and loads over 2000 Erlang, TRPO achieves a 4-times decrease in blocking concerning the best heuristic, being ideal for the future scenario of demand for connection requests~\cite{5681026}, 
highlighting the benefits of applying DRL techniques to the RMSCA problem. This improvement can be explained by the monotonic-improving behaviour of the TRPO agent~\cite{schulman2017trust}, which requires conservatively exploring the environment throughout the training to establish the confidence zone for its learning algorithm. Since its performance depends directly on the number of events that it must process during the training stage,  the more connection requests it processes - a situation that occurs at high traffic loads - the better its performance. As a result, the blocking increment performed by the agent according to the traffic load is lower than the one experienced by the heuristics. 

\section{Conclusion}
This paper presents a deep reinforcement learning approach applied for the first time in the literature to solve the routing, modulation format, spectrum and core allocation problem in dynamic multicore elastic optical networks. Simulation results show that the deep reinforcement learning approach offers a significant performance advantage over the best heuristic strategy studied. 

Further research on improving the DRL approach performance should focus on hyperparameter tuning. For instance, applying transfer learning techniques or graph neural networks to cover a broader range of topologies with decreased computational effort, increasing the size of the data to be processed to study fibers with more cores and investigating different reward schemes that differentiate the reward according to the cause of blocking (e.g. crosstalk, capacity unavailability, fragmentation or optical reach). 

Additionally, we would like to explore explainability techniques that might help understand how the agent makes its decisions to improve current heuristics. 

We expect these results and the code made available in the Git repository to help the research community study the benefits of deep reinforcement learning in the area of optical networks. 

\section*{Acknowledgements}
Financial support from projects: DI-PUCV (039.437/2020, 039.382/2021); ANID FOVI 210082; ANID FONDECYT Iniciación (11201024, 11220650, 11190710); ANID Magister Nacional (/2020-22201418, /2021-22210736), are gratefully acknowledged.

\bibliography{mybibfile}

\end{document}